\pdfoutput=1

\documentclass[11pt]{article}

\usepackage{acl}

\usepackage{times}
\usepackage{latexsym}
\usepackage{graphicx}
\usepackage{amsmath}
\usepackage{amssymb}

\usepackage[T1]{fontenc}

\usepackage[utf8]{inputenc}

\usepackage{microtype}

\title{SimpleBERT: A Pre-trained Model That Learns to Generate Simple Words}

\author{Renliang Sun \\
  Peking University \\
  \texttt{sunrenliangpku@gmail.com} \\\And
  Xiaojun Wan \\
  Peking University \\
  \texttt{wanxiaojun@pku.edu.cn} \\}

\begin{document}
\maketitle
\begin{abstract}
Pre-trained models are widely used in the tasks of natural language processing nowadays. However, in the specific field of text simplification, the research on improving pre-trained models is still blank. In this work, we propose a continued pre-training method for text simplification. Specifically, we propose a new masked language modeling (MLM) mechanism, which does not randomly mask words but only masks simple words. The new mechanism can make the model learn to generate simple words. We use a small-scale simple text dataset for continued pre-training and employ two methods to identify simple words from the texts. We choose BERT, a representative pre-trained model, and continue pre-training it using our proposed method. Finally, we obtain SimpleBERT, which surpasses BERT in both lexical simplification and sentence simplification tasks and has achieved state-of-the-art results on multiple datasets. What's more, SimpleBERT can replace BERT in existing simplification models without modification.
\end{abstract}

\section{Introduction}

The goal of text simplification is to reduce complex text to a more comprehensible text while keeping its meaning intact \cite{alva2020data, sikka2020survey}. This technology can provide convenience for children, non-native speakers, and people with dyslexia \cite{shardlow2014survey}. 

In recent years, a small number of studies have applied pre-trained models to the field of text simplification and achieved good results. \citeauthor{maruyama2019extremely} \shortcite{maruyama2019extremely} used TransformerLM to help extremely low-resource sentence simplification.
\citeauthor{qiang2020lexical} \shortcite{qiang2020lexical} proposed BERT-LS to generate simple substitutions for lexical simplification. \citeauthor{martin2020multilingual} \shortcite{martin2020multilingual} proposed unsupervised mining techniques and combined them with pre-trained models. \citeauthor{jiang2020neural} \shortcite{jiang2020neural} initialized the encoder with BERT and proposed a seq2seq model. 

However, simply applying pre-trained models to text simplification may be problematic. In the pre-training stage, the models are trained in a self-supervised way with a large number of ordinary texts, and the models are expected to learn to generate ordinary words instead of simple words. 
Besides, some tasks such as lexical simplification do not have large-scale training data. The effect of fine-tuning may not be good as expected. Therefore, it is worth investigating how to make pre-trained models learn to generate simple words.

In this work, we are committed to improving the pre-trained model like BERT \cite{devlin2019bert} to help both lexical and sentence simplification. We are inspired by the works of \citeauthor{gururangan2020don} \shortcite{gururangan2020don} and
\citeauthor{gu2020train} \shortcite{gu2020train}. They conclude that a task-guided pre-training stage using the task-related data may be helpful for downstream tasks.


Specifically, we collect a set of simple texts and then employ two methods to automatically and coarsely identify simple words in texts. We propose a new masked language modeling (MLM) mechanism which only masks the simple words in the texts and then let the model learn to predict the masked simple words. We choose a representative pre-trained model, BERT, and continue pre-training it using the method proposed above. Thus, we obtain a new pre-trained model for text simplification. We name it SimpleBERT. We experiment with SimpleBERT on several tasks of text simplification to demonstrate its strong performance and generality. 


In summary, the main contributions of our work are that (1) We propose SimpleBERT with a new MLM mechanism as the backbone for text simplification tasks. SimpleBERT can replace BERT in existing simplification models without any modification. (2) Our proposed method achieves state-of-the-art results on multiple datasets in text simplification, and it outperforms the original BERT on all datasets. (3) To our best knowledge, we are the first to explore how to improve pre-trained models in text simplification. We will release the code and model to promote future research when the paper is accepted.

\section{Continued Pre-training for Text Simplification}

We choose BERT \cite{devlin2019bert} which has been applied to text simplification tasks to verify the effectiveness of our method. We continue pre-training BERT and obtain a new model for text simplification, which we name SimpleBERT.

\subsection{SimpleBERT}

 To introduce SimpleBERT, we need to construct a set of simple texts at first. Then we need to identify the simple words in the texts\footnote{In practice, we only identify those words that contain English letters. That is to say, for invalid information such as pure numbers and punctuation, we treat them as complex words.}. The details of the simple texts and the identification methods will be described in the following subsections. 
 
 To make the pre-trained model learn to generate simple words, we propose a new masked language modeling (MLM) mechanism. For a simple word that is identified, we follow the original BERT to give a probability of 15\% that the word will be replaced with the [MASK] symbol. The following example can illustrate the new MLM mechanism:

\begin{figure}[h]
\centering
\includegraphics[width=7.5cm]{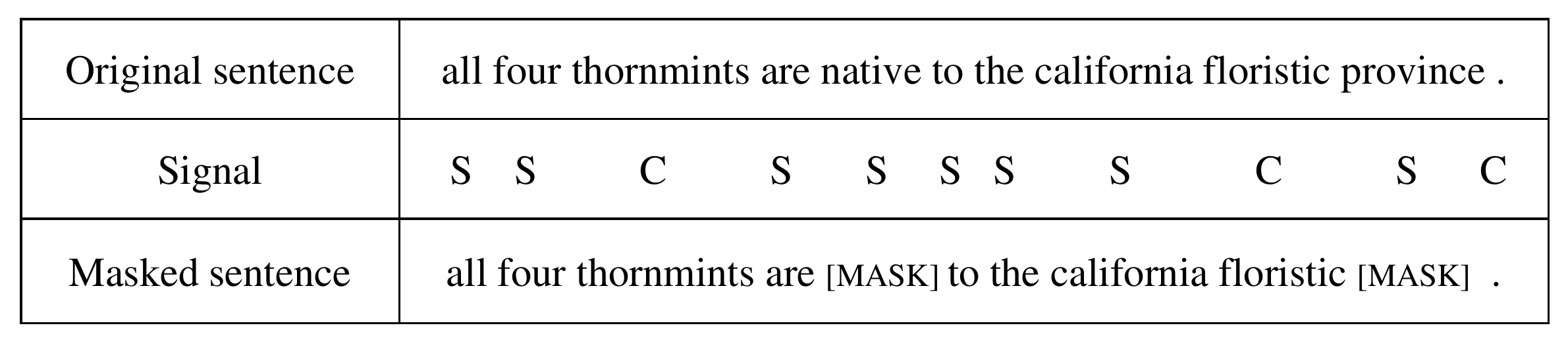} 
\caption{An illustration of the new MLM mechanism. We use S to mark the word identified as simple and C to mark the word identified as complex.}
\label{fig1}
\end{figure}




Then, the model will learn to generate the masked simple word, and the probability for generating the masked word can be denoted as:
\begin{equation}
\begin{aligned}
    P_{\theta}(x_l|x_1,\cdots,x_{l-1},x_{l+1},\cdots,x_n) \\ =  softmax(W_l^Tf_{\theta}(S\backslash l))
\end{aligned}
\end{equation}

\noindent $x$ represents a token in the text. $\theta$ represents the parameters. $l$ represents the position of the simple word being masked in the text. $n$ represents the number of words of the text. $W_l$ represents a one-hot vector for the $l$-th token. $S\backslash l$ represents a set of tokens in which the $l$-th token is replaced with [MASK]. $f_{\theta}(S\backslash l)\in \mathbb{R}^M$ represents a multi-layer bidirectional transformer model.


\subsection{Simple Text Dataset for Continued Pre-training}

We use two commonly used corpus named WikiLarge and Newsela in text simplification for continued pre-training.

WikiLarge is the largest dataset in text simplification constructed by \citeauthor{zhang2017sentence} \shortcite{zhang2017sentence}.
We use the simplified sentences in the training set as resources for continued pre-training. We also use articles in the Newsela corpus which is divided into five difficulty levels. Level 0 represents the most complicated article, and level 4 represents the simplest article. We choose English articles in the Newsela corpus for continued pre-training and limit the difficulty level of articles to 2, 3, and 4. In practice, we discard sentences with the number of words less than or equal to 5.



\subsection{Simple Words Identification}

We design two methods to identify simple words from the text. The first is based on human ratings, and the second is based on neural networks.


Human evaluation is an essential criterion for judging whether a word is simple. We use a lexicon with complexity scores proposed by \citeauthor{maddela2018word} \shortcite{maddela2018word} as the human ratings method. We select all the words with an average score of 4 points or less and build a simple word list $V_{simple}$. To determine whether a word is simple, we only need to determine whether the word is in $V_{simple}$. 




We use the SEQ model proposed by \citeauthor{gooding2019complex} \shortcite{gooding2019complex} as the neural networks method. 
SEQ makes full use of the context of the recognized words and achieves state-of-the-art results on the CWI 2018 shared task \cite{yimam2018report}. 
Given a piece of text as input, the SEQ model will automatically identify which words are simple and which words are complex.

\begin{table*}[h]
\centering
\small
\resizebox{0.8\textwidth}{!}{
\begin{tabular}{lcccccc}
\hline
\multicolumn{1}{l|}{}                      & \multicolumn{3}{c|}{LexMTurk}                                            & \multicolumn{3}{c}{BenchLS}                         \\
\multicolumn{1}{l|}{}                      & P(\%)     & R(\%)        & \multicolumn{1}{c|}{F1(\%)}            & P(\%)     & R(\%)        & F1(\%)            \\ \hline
Baselines                                  & \multicolumn{6}{l}{}                                                                                                           \\ \hline
\multicolumn{1}{l|}{Horn}                  & 15.3          & 13.4          & \multicolumn{1}{c|}{14.3}                & 23.5          & 13.1          & 16.8                \\
\multicolumn{1}{l|}{Glava\v{s}}          & 15.1          & 12.2          & \multicolumn{1}{c|}{13.5}                & 14.2          & 19.1          & 16.3                \\
\multicolumn{1}{l|}{Paetzold-CA}           & 17.7          & 14.0          & \multicolumn{1}{c|}{15.6}                & 18.0          & 25.2          & 21.0                \\
\multicolumn{1}{l|}{Paetzold-NE}           & 31.0          & 14.2          & \multicolumn{1}{c|}{19.5}                & \textbf{27.0} & 20.9          & 23.6                \\
\multicolumn{1}{l|}{BERT-LS}               & 29.6          & 23.0          & \multicolumn{1}{c|}{25.9}                & 23.6          & 32.0          & 27.2                \\
\multicolumn{1}{l|}{LSBERT(BERT)}          & 33.3          & 25.9          & \multicolumn{1}{c|}{29.1}                & 25.2          & 34.2          & 29.0                \\ \hline
Our methods                                & \multicolumn{6}{l}{}                                                                                                           \\ \hline
\multicolumn{1}{l|}{LSBERT(SimpleBERT-NN)} & 34.3          & 26.6          & \multicolumn{1}{c|}{30.0(+0.9)}          & 25.0          & 33.9          & 28.8(-0.2)          \\
\multicolumn{1}{l|}{LSBERT(SimpleBERT-HR)} & \textbf{35.3} & \textbf{27.5} & \multicolumn{1}{c|}{\textbf{30.9(+1.8)}} & 26.9          & \textbf{36.4} & \textbf{30.9(+1.9)} \\ \hline
\end{tabular}}
\caption{The results of substitution generation. We use \textbf{Bold} to mark the best results, and we report the differences between the model using SimpleBERT and the model using the original BERT. We use SimpleBERT-HR to represent the method using human ratings for simple words identification and use SimpleBERT-NN to represent the method using neural networks for simple words identification.}
\label{tab:SG}
\end{table*}

\begin{table*}[h]
\centering
\small
\resizebox{0.8\textwidth}{!}{
\begin{tabular}{lcccc}
\hline
\multicolumn{1}{l|}{}                      & \multicolumn{2}{c|}{LexMTurk}                                  & \multicolumn{2}{c}{BenchLS}      \\
\multicolumn{1}{l|}{}                      & Precision(\%)           & \multicolumn{1}{c|}{Accuracy(\%)}            & Precision(\%)  & Accuracy(\%)            \\ \hline
\multicolumn{5}{l}{Baselines}                                                                                                                  \\ \hline
\multicolumn{1}{l|}{Horn}                  & 76.1                & \multicolumn{1}{c|}{66.3}                & 54.6       & 34.1                \\
\multicolumn{1}{l|}{Glava\v{s}}          & 71.0                & \multicolumn{1}{c|}{68.2}                & 48.0       & 25.2                \\
\multicolumn{1}{l|}{Paetzold-CA}           & 57.8                & \multicolumn{1}{c|}{39.6}                & 42.3       & 42.3                \\
\multicolumn{1}{l|}{Paetzold-NE}           & 67.6                & \multicolumn{1}{c|}{67.6}                & 64.2       & 43.4                \\
\multicolumn{1}{l|}{BERT-LS}               & 77.6                & \multicolumn{1}{c|}{77.6}                & 60.7       & 60.7                \\
\multicolumn{1}{l|}{LSBERT(BERT)}          & 86.4                & \multicolumn{1}{c|}{79.2}                & 69.7       & 61.6                \\ \hline
\multicolumn{5}{l}{Our methods}                                                                                                                \\ \hline
\multicolumn{1}{l|}{LSBERT(SimpleBERT-NN)} & 86.6(+0.2)          & \multicolumn{1}{c|}{\textbf{84.0(+4.8)}} & 67.0(-2.7) & 62.9(+1.3)          \\
\multicolumn{1}{l|}{LSBERT(SimpleBERT-HR)} & \textbf{88.0(+1.6)} & \multicolumn{1}{c|}{82.0(+2.8)}          & \textbf{70.7(+1.0)} & \textbf{63.7(+2.1)} \\ \hline
\end{tabular}}
\caption{The results of full lexical simplification pipeline. }
\label{tab:pipeline}
\end{table*}

\section{Evaluation Setup}


We employ three representative tasks for experiments: substitution generation (SG), full lexical simplification (LS) pipeline, and sentence simplification (SS).
For the first two tasks, we choose to test on two commonly used datasets: LexMTurk \cite{horn2014learning} and BenchLS \cite{paetzold2016benchmarking}. For the third task, we choose to test on the high-quality Newsela dataset \cite{xu2015problems}.  More training details can be found in Appendix \ref{appendix:A}. 








\section{Evaluation Results on Substitution Generation}

The substitution generation task can be defined as: For the target complex word in an instance, the model needs to generate multiple simple candidate words. This task uses precision (P), recall (R), and F1 score (F1) to measure the quality of the generated candidate words. 


We choose several representative models as baselines, including Horn \cite{horn2014learning}, Glava\v{s} \cite{glavavs2015simplifying}, Paetzold-CA \cite{paetzold2016unsupervised}, Paetzold-NE \cite{paetzold2017lexical}, BERT-LS \cite{qiang2020lexical}, and LSBERT \cite{qiang2020lsbert}. 
The LSBERT model masks the target complex word in a sentence and then feeds it into the BERT model to generate simple candidate words. 
We use the code of the LSBERT model and replace BERT with our proposed SimpleBERT for comparison. The experimental results are shown in Table \ref{tab:SG}.

The results illustrate that the performance of SimpleBERT has surpassed the BERT model on both datasets. The F1 score of SimpleBERT on the LexMTurk dataset exceeds that of the BERT model by 1.8 points. The F1 score of SimpleBERT on the BenchLS dataset exceeds that of the BERT model by 1.9 points. In addition, LSBERT(SimpleBERT-HR) has achieved state-of-the-art results on this task, which means that SimpleBERT is adept at generating diversely simple words. We select example outputs to illustrate the effectiveness of SimpleBERT , as shown in Table \ref{tab:example_LS}. 

\section{Evaluation Results on Full Lexical Simplification Pipeline}

The task of full lexical simplification (LS) pipeline can be defined as follows: for the target complex word in an instance, the model needs to replace it with one simple word. The evaluation measures for the full LS pipeline are precision and accuracy. The baselines we use are the same as those for Task 1. We also use the LSBERT model for experiments. We keep the substitution ranking rules the same and only replace BERT with SimpleBERT for comparison. The results are shown in Table \ref{tab:pipeline}.

SimpleBERT performs well on this task. For the LexMTurk dataset, SimpleBERT exceeds the BERT model by 1.6 points in precision and 4.8 points in accuracy. For the BenchLS dataset, SimpleBERT exceeds the BERT model by 1 point in precision and 2.1 points in accuracy. SimpleBERT has achieved state-of-the-art results on both datasets, illustrating that in combination with the existing ranking rules, SimpleBERT can generate the most appropriate simple substitutions. We also give example outputs, as shown in Table \ref{tab:example_LS_pipeline}.

\section{Evaluation Results on Sentence Simplification}
\subsection{Automatic Evaluation Results}
Sentence simplification converts complex sentences into simple sentences while keeping the meanings unchanged. We use SARI \cite{xu2016optimizing} and FKGL \cite{kincaid1975derivation} to evaluate the sentence simplification task.



We choose several representative models as baselines, including PBMT-R \cite{wubben2012sentence}, Hybrid \cite{narayan2014hybrid}, DRESS \cite{zhang2017sentence}, DRESS-LS \cite{zhang2017sentence}, EditNTS \cite{dong2019editnts} and Transformer(BERT) \cite{jiang2020neural}. The Transformer(BERT) model has a standard encoder-decoder framework. The encoder uses BERT for parameters initialization, and parameters of the decoder are initialized randomly. We have implemented the Transformer(BERT) model and only use SimpleBERT to replace BERT in the encoder.

\begin{table}[h]
\centering
\resizebox{0.48\textwidth}{!}{
\begin{tabular}{lcc}
\hline
\multicolumn{1}{l|}{}               & SARI$\uparrow$           & FKGL$\downarrow$          \\ \hline
\multicolumn{1}{l|}{PBMT-R}         & 26.24          & 8.13          \\
\multicolumn{1}{l|}{Hybrid}         & 34.73          & 4.52          \\
\multicolumn{1}{l|}{DRESS}          & 38.37          & 4.65          \\
\multicolumn{1}{l|}{DRESS-LS}       & 38.04          & 4.78          \\
\multicolumn{1}{l|}{EditNTS}        & 39.28          & \textbf{3.80} \\
\multicolumn{1}{l|}{Transformer(BERT)}    & 39.59          & 4.07          \\ \hline
\multicolumn{3}{l}{Our methods}                                      \\ \hline
\multicolumn{1}{l|}{Transformer(SimpleBERT-NN)} & \textbf{40.17(+0.58)}         & 3.88(-0.19)          \\
\multicolumn{1}{l|}{Transformer(SimpleBERT-HR)} & 39.70(+0.11) & 3.85(-0.22)          \\ \hline
\end{tabular}}
\caption{The results of sentence simplification on the Newsela dataset.}
\label{tab:SS}
\end{table}

As shown in Table \ref{tab:SS}, the SimpleBERT outperforms the BERT model by 0.58 points in SARI and by 0.22 points in FKGL. In addition, Transformer(SimpleBERT-NN) outperforms the EditNTS model by 0.89 points in SARI. Thus, Transformer(SimpleBERT-NN) achieves state-of-the-art results on the Newsela dataset, illustrating that SimpleBERT can contribute to sentence simplification, even though it is trained at the word level. We show example outputs in Table \ref{tab:example_SS}.

\subsection{Human Evaluation Results}

To explore how SimpleBERT helps sentence simplification, we hire three workers to perform a human evaluation of the output of different models. We randomly select 100 complex-simple sentence pairs from the test set of the Newsela dataset. Following \citeauthor{dong2019editnts} \shortcite{dong2019editnts}, workers are required to rate sentences using a five-point Likert scale on Simplicity (S), Fluency (F), and Adequacy (A). Following \citeauthor{xu2016optimizing} \shortcite{xu2016optimizing}, we use Simplicity Gain (S+) to indicate how good the lexical simplification is in a sentence-level simplification. 



    
    
    
    

\begin{table}[h]
\centering
\resizebox{0.48\textwidth}{!}{
\begin{tabular}{l|cccc}
\hline
                       & S             & F             & A             & S+             \\ \hline
EditNTS                & 2.91          & 3.82          & 2.64          & 9.33           \\
Transformer(BERT)          & 2.92          & 4.51          & 3.44          & 18.67          \\
Transformer(SimpleBERT-NN) & \textbf{3.02} & \textbf{4.56} & 3.42          & \textbf{20.67} \\
Transformer(SimpleBERT-HR) & 3.01          & 4.52          & \textbf{3.48} & 19.00          \\ \hline
\end{tabular}}
\caption{The results of human evaluation on the Newsela dataset. We use \textbf{Bold} to mark the best results. Transformer(SimpleBERT-NN) achieves best results in the dimension of Simplicity, which is consistent with the results of the automatic evaluation.}
\label{tab:human_evaluation}
\end{table}


From Table \ref{tab:human_evaluation}, we can draw two conclusions: (1) SimpleBERT is more capable than BERT in generating appropriate simple words. (2) Reducing the complexity from the word level can greatly help sentence simplification.

\section{Additional Analysis}

We perform ablation experiments to explore the effect of the constructed simple text and the new MLM mechanism. We also made a comparison of the two methods for identifying simple words, namely human ratings and neural networks. However, due to space limitations, we have to put the experimental details into Appendix \ref{appendix:C}. When the paper is accepted, we will put the two experiments into the additional page of the  main text.

\section{Conclusion}

In this paper, we propose SimpleBERT by continually pre-training BERT on a simple text dataset with a new masked language modeling mechanism. SimpleBERT can be integrated directly into existing models without any code modifications, demonstrating great versatility. More importantly, SimpleBERT has surpassed the original BERT in lexical and sentence simplification tasks and achieves state-of-the-art results on multiple datasets. 

\section{Ethical Consideration}

We use the text of the WikiLarge dataset and the Newsela corpus for continue pre-training. The WikiLarge dataset is under the MIT License. For the Newsela corpus, it is only available to academic partners and we have the permission to use it.

We recruit three workers on a first-come, first-served basis after posting the human evaluation tasks.
We pay each worker 30 US dollars, that is, 0.3 dollars for scoring a complex-simple sentence pairs. And we let them complete the evaluation within 72 hours.

\bibliography{custom}
\bibliographystyle{acl_natbib}

\appendix

\section{Training Details}
\label{appendix:A}

For the first two tasks, we choose to test on two commonly used datasets: LexMTurk and BenchLS. LexMTurk contains 500 instances, of which fifty native English speakers suggest the simpler substitutions \cite{horn2014learning}.  BenchLS contains 929 instances. It is a combination of LexMTurk and another dataset named LSeval after correcting errors \cite{paetzold2016benchmarking}. Both of them only have test set and do not include training set.

For the third task, we choose to test on the Newsela dataset. Its test set has 1,077 instances. Previous studies have shown that the Newsela dataset is of higher quality than the Wikipedia-based dataset \cite{xu2015problems}.

For lexical simplification, previous studies used different versions of BERT for different tasks to achieve better results \cite{qiang2020lsbert}. We follow their setup for a fair comparison. For the substitution generation task, we use BERT-Large for continued pre-training. For the full lexical simplification pipeline task, we use the BERT-Large-WWM for continued pre-training. For the sentence simplification task, we use the code from \citeauthor{liu2019text} \shortcite{liu2019text} to implement the model and get the same results with \citeauthor{jiang2020neural} \shortcite{jiang2020neural}. 

We continue training BERT on 4 GTX 1080ti. The batch size on each GPU is 4, and the gradient accumulation value is 4. Thus, the total batch size for training is 64. For the tasks of lexical simplification and sentence simplification, we set the max sequence length to 256 and 512, respectively. We follow \cite{gururangan2020don} and report more hyperparameters in Table \ref{tab:hyperparams}.

\begin{table}[h]
\centering
\begin{tabular}{cc}
\hline
Hyperparameter              & Assignment \\ \hline
Number of training epochs   & 10         \\
Batchsize                   & 4          \\
Gradient accumulation value & 4          \\
Learning rate               & 5e-5       \\
Learning rate optimizer     & Adam       \\
Adam epsilon                & 1e-8       \\
Adam beta weights           & 0.9 0.999  \\ \hline
\end{tabular}
\caption{Hyperparameters for continue pre-training}
\label{tab:hyperparams}
\end{table}

Regarding the text used for continued pre-training, we use the simple text dataset constructed by WikiLarge and Newsela introduced before. We use programs to compare the constructed text data with the test set of each task, and then remove the same sentences in the text data to avoid overlapping. To avoid SimpleBERT learning information from the test set of Newsela, we only use the training set of WikiLarge for continued pre-training in sentence simplification. We then use the training set of the Newsela dataset to fine-tune the whole model.

For the third task of sentence simplification, we use EASSE \cite{alva2019easse} to obtain the SARI values and FKGL values.
Previous works used different scripts for evaluation. For a fair comparison, we use EASSE to evaluate the outputs of all the simplification systems.

\section{Additional Analysis}
\label{appendix:C}

\subsection{Ablation Study}

To explore the effect of the constructed simple text and the new MLM mechanism, we design the following ablation experiments on the substitution generation task. We still use LSBERT and use the following variants of pre-trained models in the method for comparison. 

\begin{itemize}

    \item \textbf{BERT:} It uses the original BERT.
    
    \item \textbf{BERT-C:} It uses the simple text dataset constructed for continued pre-training on BERT. Nevertheless, the MLM mechanism is the same as the original BERT, which randomly masks words.
    
    \item \textbf{BERT-M:} It uses ordinary text\footnote{The source of the ordinary text is the original sentences in the training set of the WikiLarge dataset and the level 0 articles of the Newsela corpus. The scale of ordinary text data is roughly the same as that of simple text dataset.} for continued pre-training on BERT. The human rating method is used to identify simple words, and our new MLM mechanism is used to mask simple words.
    
    \item \textbf{SimpleBERT:} It uses the simple text dataset we constructed for continued pre-training on BERT. The human rating method is used to identify simple words, and our new MLM mechanism is used to mask simple words.

\end{itemize}

The results of ablation experiments are shown in Table \ref{tab:ablation}. From the results of BERT-C, it can be seen that using the constructed simple text for continued pre-training has improved the results on both datasets. From the results of BERT-M, it can be seen that using our MLM mechanism to mask simple words can improve the results on both datasets even if the ordinary text is used. The results of SimpleBERT are better than those of BERT-C and BERT-M, indicating that using the new MLM mechanism enables the model to better learn to generate simple words from the constructed text.

\begin{table}[h]
\centering
\resizebox{0.4\textwidth}{!}{
\begin{tabular}{l|cc}
\hline
           & LexMTurk   & BenchLS    \\ \hline
BERT       & 29.1       & 29.0       \\
BERT-C     & 29.8(+0.7) & 30.4(+1.4) \\
BERT-M     & 30.0(+0.9) & 30.2(+1.2) \\
SimpleBERT & 30.9(+1.8) & 30.9(+1.9) \\ \hline
\end{tabular}}
\caption{F1 scores (\%) of the pre-trained models used for ablation study on both datasets. We report the differences between the variant models and the original BERT.}
\label{tab:ablation}
\end{table}

\subsection{Human Ratings vs. Neural Networks for Simple Words Identification}

In this work, we use human ratings and neural networks to identify simple words, respectively. However, the performance of SimpleBERT obtained by using the two methods on the three tasks is different. SimpleBERT-HR performs better than SimpleBERT-NN on two lexical simplification tasks, Task 1 and Task 2. However, the opposite is true on the sentence simplification task, Task 3.

For a reasonable explanation, we select 100 instances containing a total of 2,069 words from the constructed simple text dataset and use the two methods to identify simple words separately. We make a count and find that there are 302 words for which the two methods give different identification results. Among them, there are 245 words identified as simple by human ratings while identified as complex by neural networks. The remaining 57 words are identified as simple by neural networks while identified as complex by human ratings.

That is to say, for words that are difficult to identify as complex or simple, SimpleBERT-HR tends to regard them as simple. It will enable SimpleBERT-HR to generate a relatively wider variety of candidate words for re-ranking in lexical simplification, which may help to improve the results. In contrast, sentence simplification only requires generating one simplified sentence that is less tolerant of complex words, which may account for the better performance of SimpleBERT-NN.

\section{Example Outputs of the Three Tasks}
\label{appendix:B}

\begin{table*}[h!]
\centering
\resizebox{16.5cm}{!}{
\begin{tabular}{ll}
\hline
\multicolumn{1}{|l|}{Original sentence} & \multicolumn{1}{l|}{\begin{tabular}[c]{@{}l@{}}the term is also at times used as a \underline{derogatory} term for films or sequels of films that are of inferior\\ quality , or are not expected to find financial success .\end{tabular}}                          \\ \hline
\multicolumn{1}{|l|}{Gold standard}     & \multicolumn{1}{l|}{foul, bad, different, mean, degrading, insulting, negative, demeaning, offensive, rude, pejorative}                                                                                                                    \\
\multicolumn{1}{|l|}{LSBERT(BERT)}              & \multicolumn{1}{l|}{slang, loose, sarcastic, descriptive, positive, humorous, neutral, generic, \textbf{negative}, defensive}                                                                                             \\
\multicolumn{1}{|l|}{LSBERT(SimpleBERT-HR)}        & \multicolumn{1}{l|}{\textbf{offensive}, \textbf{bad}, \textbf{negative}, dirty, descriptive, general, \textbf{rude}, positive, racial, \textbf{mean}} \\ \hline
\end{tabular}}
\caption{Example outputs of LSBERT(SimpleBERT-HR) and LSBERT(BERT) for substitution generation. We \underline{underline} the target complex word in the original sentence, and use \textbf{bold} to mark the candidate words appearing in the gold standard. BERT generates many complex words such as ``sarcastic'' and ``generic''. The candidate words generated by SimpleBERT are generally simpler than those generated by BERT and appear more often in the gold standard.}
\label{tab:example_LS}
\end{table*}

\begin{table*}[h!]
\centering
\resizebox{16.5cm}{!}{
\begin{tabular}{|l|l|}
\hline
Original sentence     & \begin{tabular}[c]{@{}l@{}}the united states \underline{convened} a 13 - nation conference of the international opium commission in 1909\\ in shanghai , china in response to increasing criticism of the opium trade .\end{tabular} \\ \hline
Gold standard         & \begin{tabular}[c]{@{}l@{}}opened, started, assembled, called, gathered, created, made, aggregated, brought together, hosted, \\ organized, \textbf{held}, met, ordered\end{tabular}                                              \\
LSBERT(BERT)          & summoned                                                                                                                                                                                                                 \\
LSBERT(SimpleBERT-HR) & \textbf{held}                                                                                                                                                                                                                     \\ \hline
\end{tabular}}
\caption{Example outputs of LSBERT(BERT) and LSBERT(SimpleBERT-HR) for full lexical simplification pipeline. Although both BERT and SimpleBERT can generate words with the same meaning as the original word, it is apparent that SimpleBERT generates ``held'' which is simpler than BERT's ``summoned''.}
\label{tab:example_LS_pipeline}
\end{table*}

\begin{table*}[h!]
\centering
\resizebox{16.5cm}{!}{
\begin{tabular}{ll}
\hline
\multicolumn{1}{|l|}{Original sentence}  & \multicolumn{1}{l|}{\begin{tabular}[c]{@{}l@{}}in minnesota , youth hockey has long had a reputation for competitiveness , but now , youth basketball\\ might be catching up .\end{tabular}}                                          \\ \hline
\multicolumn{1}{|l|}{Reference sentence} & \multicolumn{1}{l|}{minnesota is famous for competitive youth hockey .}                                                                                                                       \\
\multicolumn{1}{|l|}{Transformer(BERT)}               & \multicolumn{1}{l|}{in minnesota , youth hockey has long been reputation for competitiveness .}                                                                                               \\
\multicolumn{1}{|l|}{Transformer(SimpleBERT-NN)}         & \multicolumn{1}{l|}{in minnesota , youth hockey has long been known for competitiveness .}                                                                                                    \\ \hline
                                         &                            \\  
\hline
\multicolumn{1}{|l|}{Original sentence}  & \multicolumn{1}{l|}{\begin{tabular}[c]{@{}l@{}}the program tries to reverse this trend by giving students simple and fun ways to include others during\\ lunchtime -- making sure no one eats alone , said laura talmus .\end{tabular}} \\ \hline
\multicolumn{1}{|l|}{Reference sentence} & \multicolumn{1}{l|}{it gives students fun ways to include others during lunchtime .}                                                                                                          \\
\multicolumn{1}{|l|}{Transformer(BERT)}               & \multicolumn{1}{l|}{\begin{tabular}[c]{@{}l@{}}the program tries to reverse this trend by giving students simple and fun ways to include others during\\ lunchtime .\end{tabular}}                                                     \\
\multicolumn{1}{|l|}{Transformer(SimpleBERT-NN)}         & \multicolumn{1}{l|}{it gives students simple and fun ways to include others during lunchtime .}                                                                                               \\ \hline
\end{tabular}}
\caption{Example outputs of Transformer(BERT) and Transformer(SimpleBERT-NN) for sentence simplification. In the first example, BERT generates ``reputation'' while SimpleBERT generates ``known'' which has the same meaning but is simpler than ``reputation''. In the second example, SimpleBERT removes the complex words ``reserve this trend'' from the original sentence and generates a sentence that is much closer to the reference.}
\label{tab:example_SS}
\end{table*}



\end{document}